%% file: aaai25.tex
\documentclass[letterpaper]{article} 
\usepackage{aaai25}  
\usepackage{times}  
\usepackage{helvet}  
\usepackage{courier}  
\usepackage[hyphens]{url}  
\usepackage{graphicx} 
\def\UrlFont{\rm}  
\usepackage{natbib}  
\usepackage{caption} 
\usepackage{algorithm}
\usepackage{algorithmic}

\usepackage{amsmath}
\usepackage{multirow}
\usepackage{subcaption}

\usepackage[utf8]{inputenc} 
\usepackage[T1]{fontenc}    
\usepackage{booktabs}       
\usepackage{amsfonts}       
\usepackage{nicefrac}       
\usepackage{microtype}      
\usepackage{xcolor}         

\usepackage{newfloat}
\usepackage{listings}
\DeclareCaptionStyle{ruled}{labelfont=normalfont,labelsep=colon,strut=off} 
\floatstyle{ruled}
\newfloat{listing}{tb}{lst}{}
\floatname{listing}{Listing}
\title{DC-PCN: Point Cloud Completion Network with\\ Dual-Codebook Guided Quantization}
\author {
    Qiuxia Wu\textsuperscript{\rm 1},
    Haiyang Huang\textsuperscript{\rm 1},
    Kunming Su\textsuperscript{\rm 1},
    Zhiyong Wang\textsuperscript{\rm 2},
    Kun Hu\textsuperscript{\rm 2}\thanks{Corresponding author}
}
\affiliations {
    \textsuperscript{\rm 1}South China University of Technology\\
    \textsuperscript{\rm 2}The University of Sydney\\
    qxwu@scut.edu.cn, \{HaiYangHuang.5461, kunmingsu07\}@gmail.com, 
    \{zhiyong.wang, kun.hu\}@sydney.edu.au
}
\usepackage{bibentry}
\begin{document}
\maketitle
\begin{abstract}
Point cloud completion aims to reconstruct complete 3D shapes from partial 3D point clouds. With advancements in deep learning techniques, various methods for point cloud completion have been developed. Despite achieving encouraging results, a significant issue remains: these methods often overlook the variability in point clouds sampled from a single 3D object surface. This variability can lead to ambiguity and hinder the achievement of more precise completion results. 
Therefore, in this study, we introduce a novel point cloud completion network, namely Dual-Codebook Point Completion Network (DC-PCN), following an encder-decoder pipeline. The primary objective of DC-PCN is to formulate a singular representation of sampled point clouds originating from the same 3D surface. 
 DC-PCN introduces a dual-codebook design to quantize point-cloud representations from a multilevel perspective. It consists of an encoder-codebook and a decoder-codebook,  designed to capture distinct point cloud patterns at shallow and deep levels. Additionally, to enhance the information flow between these two codebooks, we devise an information exchange mechanism. This approach ensures that crucial features and patterns from both shallow and deep levels are effectively utilized for completion. Extensive experiments on the PCN, ShapeNet\_Part, and ShapeNet34 datasets demonstrate the state-of-the-art performance of our method. 
\end{abstract}
\begin{links}
\link{Code}{https://github.com/tthrvfd/dcpcn.}
\end{links}
%
\section{Introduction}
\label{sec:intro}
With the increasing popularity of 3D scanning devices such as LiDAR scanners, laser scanners, and RGB-D cameras \cite{fei2022comprehensive}, the realm of vision and robotics has witnessed a significant surge in interest in utilizing 3D data. 
Among the various formats available, point clouds have emerged as a prominent choice due to their advantages, including low memory usage and detailed 3D shape information \cite{qi2017pointnet,yuan2018pcn,mo2025motion}. 
However, raw point cloud data obtained directly from 3D sensors often suffers from incompleteness due to the limitations such as viewpoint occlusion and the resolution of 3D scanners \cite{fei2022vq}. 
This incompleteness poses significant challenges for downstream tasks, such as point cloud classification and segmentation, hindering the achievement of desired results \cite{tchapmi2019topnet, xie2020grnet, yuan2018pcn}.
\begin{figure}
    \centering
    \includegraphics[width=\linewidth]{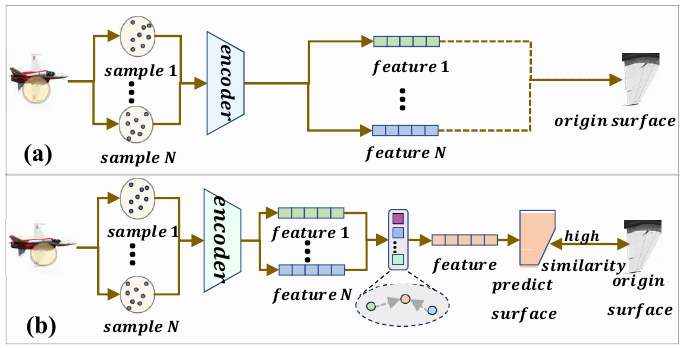}
    \caption{
    Comparison between different approaches: (a) point-based approach and (b) ours. 
    }
    \label{fig:CBvsVX}
\end{figure}
 
Thanks to advances in deep learning, a variety of point cloud completion methods have been explored. They can be broadly categorized into the \textit{voxel-based approach}  and the \textit{point-based approach} . 
The voxel-based approach \cite{choy20163d, girdhar2016learning, dai2017shape, han2017high, wu20153d} introduces a voxel paradigm to formulate structural representations for 3D objects with 3D convolutions. 
The major issue of this approach is to control the voxel granularity. 
Smaller voxel sizes typically demand high computational costs; while larger voxels can lead to substantial degradation of the structural information and the precision of the completion outcomes. 
The point-based approach has been proposed to take point cloud inputs directly such as PCN \cite{yuan2018pcn}, PF-Net \cite{huang2020pf}, MSN \cite{liu2020morphing}, TopNet \cite{tchapmi2019topnet} and GRNet \cite{xie2020grnet}. 
Recently, to enhance the extraction of intricate structural patterns, Point-Transformer \cite{zhao2021point}, Pointformer \cite{pan20213d}, and PCT \cite{guo2021pct} integrate Transformer architectures into point cloud processing  to capture long-range point dependencies. 
Nonetheless, these methods overlook the fact that point clouds sampled from a 3D object's surface can vary, leading to ambiguity issues for an identical 3D surface and ultimately incur sub-optimal completion outcomes, as illustrated in Figure~\ref{fig:CBvsVX} (a). 

Therefore, in this study, we propose a novel point cloud completion network, namely Dual-Codebook Point Completion Network (DC-PCN) following an encoder-decoder scheme. It aims to preserve as much information as possible from the point cloud while also concentrating features from the same surface into a unique, consistent representation, thereby reducing potential ambiguities in learning.
Under the inspiration of ShapeFormer \cite{yan2022shapeformer} and AutoSDF \cite{mittal2022autosdf}, DC-PCN introduces a dual-codebook design for consistent point cloud representations from a multi-level perspective. 
As illustrated in Figure~\ref{fig:CBvsVX} (b), 
the latent feature representation vector is quantized to reduce the variance caused by the point cloud sampling procedure. Specifically, the dual-codebook consists of an encoder-codebook and a decoder-codebook to explore the shallow and the deep level point cloud patterns. Moreover, a quantized information exchanging mechanism is further devised to connect and interact between the two codebooks, aiming to enhance the structural and finer detailed point cloud patterns jointly. 
Comprehensive experiments conducted on three benchmark datasets, including PCN, ShapeNet\_Part, and ShapeNet34, demonstrate that our DC-PCN achieves the state-of-the-art performance in point cloud completion. 

To summarize, our contributions are as follows:
\begin{itemize}
	\item We propose a novel codebook-based point cloud completion network, namely DC-PCN, with a dual-codebook design for consistent point cloud representations from a multi-level perspective.
	\item  A quantized information exchanging mechanism is devised to connect and interact between the two codebooks, to enhance structural and finer detailed patterns. 
	\item Comprehensive experiments on benchmark datasets including PCN, ShapeNet\_Part, and ShapeNet34 demonstrate the effectiveness of our DC-PCN.
\end{itemize}
\section{Related Work}
\paragraph{Voxel-Based Approach.} In addressing the challenges of the unstructured nature of point clouds, \cite{choy20163d, girdhar2016learning, dai2017shape, han2017high, wu20153d} introduced the voxel paradigm to formulate structured representations for 3D objects. These methods have achieved encouraging performance in point cloud completion, using the 3D convolutions to capture the spatial relationships within the voxelized data. Note that the point clouds captured from the same surface can achieve the same voxel structure if the voxel size is set appropriately, resulting in a consistent surface representation.
Nonetheless, a prevalent challenge encountered by these methods is the intricate balance strategy for controlling the voxel granularity. 
A smaller voxel size demands higher computational costs; while 
a larger voxel size typically results in substantial degradation of the structural information, reducing the precision of point cloud completion.

\paragraph{Point-Based Approach.} To address the issues of voxel-based methods, point-based studies have been proposed to take point cloud inputs directly. The pioneer study PCN \cite{yuan2018pcn} introduced a point-based point cloud completion neural network. Subsequently, various point cloud completion methods have emerged, such as PF-Net \cite{huang2020pf}, MSN \cite{liu2020morphing}, TopNet \cite{tchapmi2019topnet}, GRNet \cite{xie2020grnet}, P2c\cite{cui2023p2c}. 
Recent years, to further augment the extraction of intricate structural patterns, Point-Transformer \cite{zhao2021point}, Pointformer \cite{pan20213d}, and PCT \cite{guo2021pct} were among the initial attempts to integrate Transformer architectures into point cloud processing. SVDFormer \cite{zhu2023svdformer} proposed novel method that leverages multiple-view depth image information to observe incomplete self-shape and generate a compact global shape through Transformer architecture. Leveraging the capability of Transformer to capture long-range spatial dependencies, these methods have achieved refined results for point cloud completion. 
However, these point-based methods overlook that point clouds sampled from a 3D object surface can vary. It leads to the ambiguity challenges to obtain identical representations for a 3D surface, which ultimately incur sub-optimal completion outcomes. 
\begin{figure*}
    \centering
    \includegraphics[width=\linewidth]{ 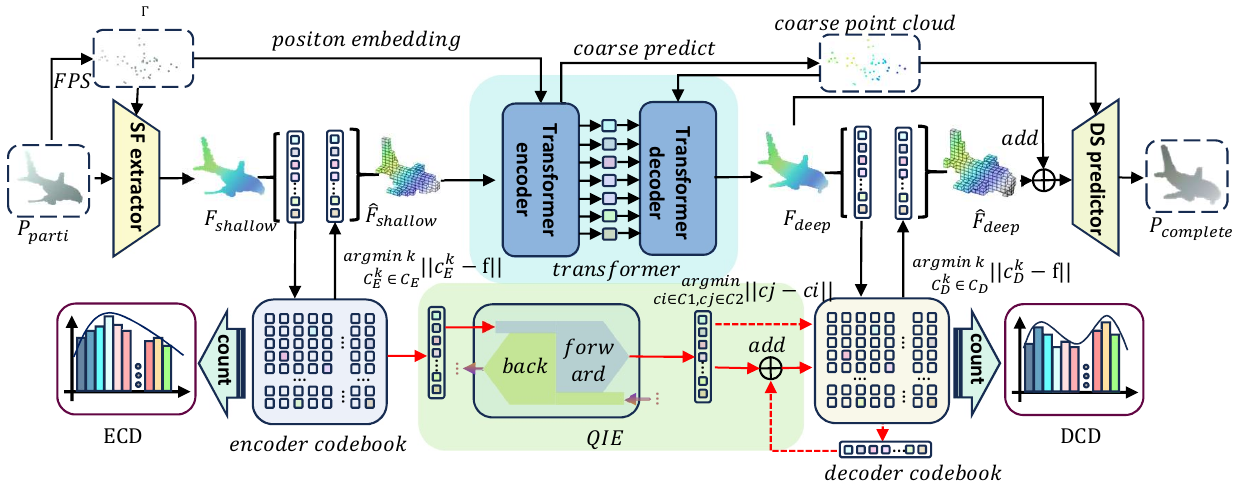}
    \caption{
    DC-PCN architecture. To achieve consistent and unambiguous latent representations for point clouds, a dual-codebook design is introduced to quantize features in both shallow and deep level, which promotes to obtain  completions in high-quality.
    }
    \label{fig:architecture}
\end{figure*}

In this study, our goal is to preserve as much information as possible from the point cloud while also concentrating features from the same surface into a unique, consistent representation, thereby reducing potential ambiguities in learning.

\section{Methodology}
\subsection{Overview of DC-PCN \& Problem Formulation}
The proposed DC-PCN follows a transformer-based encoder-decoder architecture, as illustrated in Figure~\ref{fig:architecture}. It treats point cloud completion by mapping a partial point cloud $\mathcal{P}_\text{partial} \in \mathbb{R}^{N_\text{partial}\times 3}$ to its complete form $\mathcal{P}_\text{complete} \in \mathbb{R}^{N_\text{complete}\times 3}$, which is an estimation of the ground truth $\mathcal{P}_\text{GT}$. $N_\text{partial}$ and $N_\text{complete}$ denote the number of points in the partial point cloud and complete point cloud, respectively; and 3 indicates the 3-dimensional coordinates of points. 
Note that we adopt a region-based modelling strategy, where regions are obtained by a farthest point sampling (FPS) algorithm from $\mathcal{P}_\text{partial}$.
These regions can be defined as a set of circles, each containing a subset of points in $\mathcal{P}_\text{partial}$. 
In total, we have $M$ regions with their centers $\mathbf{\Gamma}=(\mathbf{\gamma}_1,...,\mathbf{\gamma}_m,...,\mathbf{\gamma}_M)^\top$. 

As in the DC-PCN pipeline, a shallow feature extractor, noted as SF extractor, first formulates a set of shallow feature vectors: $\mathbf{F}_\text{shallow}=(\mathbf{f}_1,...,\mathbf{f}_m,...,\mathbf{f}_M)^\top$, in line with the $M$ regions. 
$\mathbf{F}_\text{shallow}$ is then used to predict a coarse point cloud,  
denoted as $\mathcal{P}_\text{coarse} \in \mathbb{R}^{N_\text{coarse} \times 3}$, where $N_\text{coarse}$ is the number of points in $\mathcal{P}_\text{coarse}$. 
$\mathcal{P}_\text{coarse}$ is used to formulate the deep features $\mathbf{F}_\text{deep} \in \mathbb{R}^{H \times C}$, where $H$ is the number of deep feature vectors and $C$ denotes the feature dimension. 
Finally, based on $\mathbf{F}_\text{deep}$, the complete point cloud $\mathcal{P}_\text{complete}$ can be obtained with the help of detailed shape predictor which is a foldingNet based module, noted as DS predictor. 

To assist this shallow-level and deep-level representation learning, our DC-PCN designs a dual-codebook quantization scheme to achieve consistent and unambiguous latent representations for point clouds. In detail, 
$\mathbf{F}_\text{shallow}$ and $\mathbf{F}_\text{deep}$ are quantized using an encoder-codebook and a decoder-codebook, respectively. Each point feature vector is represented by the closest code vector in the codebook, resulting in their discrete versions $\hat{\mathbf{F}}_\text{shallow}$ and $\hat{\mathbf{F}}_\text{deep}$.
These are combined with $\mathbf{F}_\text{deep}$ and fed to the shape predictor. 
To enable the two codebooks to work together and overcome the difference of data distribution between encoder codebook, noted as ECD and decoder codebook, noted as DCD, a quantized information exchange mechanism is devised, promoting the exchange of information between the shallow-layer and deep-layer codebooks. In the following sections, we will elaborate on the details of DC-PCN.

\subsection{Dual-Codebook for Discrete  Representations}
To maintain accurate structural and consistent information, we devise an innovative dual-codebook approach inspired by the discretization process of vector quantized variational autoencoders (VQ-VAEs) \cite{van2017neural}. This is based on the rationale that point clouds sampled from the same surface are visually similar, and their extracted features should also reflect this similarity. This discretization operation projects identical or similar surface features into the same code vector, reducing the impacts of random sampling variations in the latent space and achieving a consistent representation of point cloud samples from a 3D object.

The dual-codebook design consists of an encoder-codebook $\mathbf{C}_{E}$ and a decoder-codebook $\mathbf{C}_{D}$. 
The design places $\mathbf{C}_{E}$ prior to the transformer encoder to ensure the precise understanding of input geometric structures, while $\mathbf{C}_{D}$ is positioned after the transformer decoder to provide guidance for the output.
In detail, we have $\mathbf{C}_{E}=(\mathbf{c}_E^1,...,\mathbf{c}_E^k,,...,\mathbf{c}_E^K)^\top \in \mathbb{R}^{K \times R}$ and $\mathbf{C}_{D}=(\mathbf{c}_D^1,...,\mathbf{c}_D^k,...,\mathbf{c}_D^K)^\top \in \mathbb{R}^{K \times R}$. The two codebooks are with a consistent dimension, where $K$ and $R$ indicate the size of a codebook and the dimension of a code vector, respectively.  
Taking $\mathbf{C}_{E}$ as an example, it aims to quantize $\mathbf{f}_m$ with a posterior distribution $q(\mathbf{z}_m|\mathbf{f}_m)$ as follows:
\begin{equation}
q(\dot{\mathbf{z}}_m=\mathbf{c}_E^{k^*}|\mathbf{f}_m)=\begin{cases}
        1, \quad \mathrm{for}\,\, k^*=\mathrm{argmin}_{k}||\mathbf{f}_m-\mathbf{c}_E^k||_{2}, \\
        0, \quad  \mathrm{otherwise}.   
    \end{cases}
\end{equation}
To this end, a set of discrete representations is obtained, denoted as $\dot{\mathbf{Z}} = (\dot{\mathbf{z}}_1,...,\dot{\mathbf{z}}_m,...,\dot{\mathbf{z}}_M)^\top$. 
During training, $\mathbf{c}_E^k$ is updated in an interactive manner with $\mathbf{F}_\text{shallow}$ to adjust its distribution and align with the non-discrete representations. 
Subsequently, $\dot{\mathbf{Z}}$ is used as a proxy for $\mathbf{F}_\text{shallow}$ and is sent to the transformer encoder. 
Similarly, we denote the discrete representation set from the decoder-codebook as $\ddot{\mathbf{Z}} = (\ddot{\mathbf{z}}_1,...,\ddot{\mathbf{z}}_n,...,\ddot{\mathbf{z}}_N)^\top$.

\subsection{Quantized Information Exchanging}
\begin{figure}[htbp]
	\centering
	\begin{subfigure}{0.45\linewidth}
		\centering
		\includegraphics[width=0.9\linewidth]{ 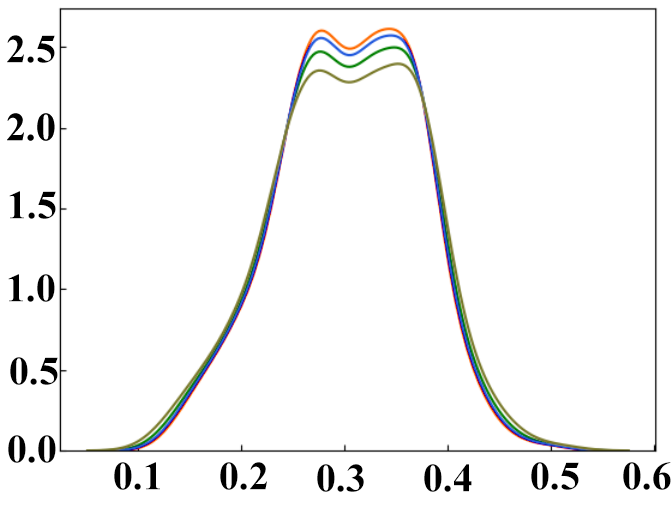}
        \caption{}
        \label{fig:C1D}
	\end{subfigure}
	\begin{subfigure}{0.45\linewidth}
		\centering
		\includegraphics[width=0.9\linewidth]{ 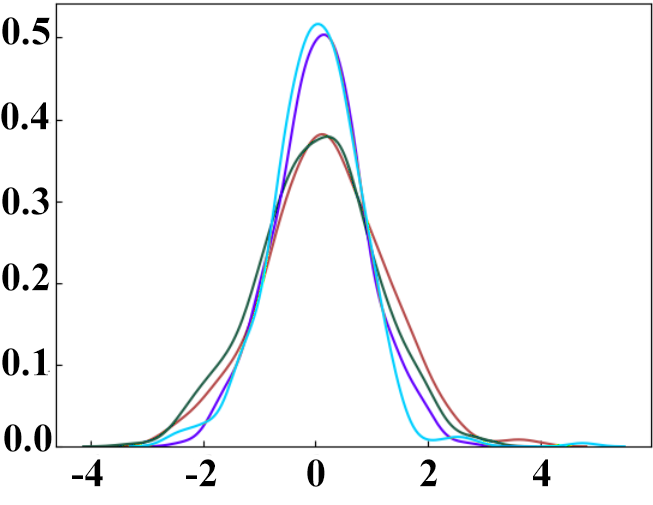}
        \caption{}
        \label{fig:C2D}
	\end{subfigure}
 
    \caption{Visualization of codebook distributions of (a) the encoder-codebook and (b) and the decoder-codebook, where an obvious discrepancy between them can be observed. 
    }
    \label{fig:CD}
\end{figure}

The two codebooks characterize shallow and deep point cloud feature distributions, and the quantized point cloud representations $\dot{\mathbf{Z}}$ and $\ddot{\mathbf{Z}}$ each contain unique patterns that, 
when combined, are assumed to achieve both structural and finer detailed enhancements for point cloud completion. 
Hence, we propose facilitating the incorporation between $\dot{\mathbf{Z}}$ and $\ddot{\mathbf{Z}}$ with a quantized information exchanging mechanism, noted as QIE. 
This mechanism comprises three components: \textit{code deduplication}, \textit{code distribution re-targeting} , and \textit{code merging }. The details are discussed as follows.
\input{ tables/PCN}
\paragraph{Code Deduplication.} In pursuit of a high information density and representativeness to better captures the intrinsic structral pattern, we propose to deduplicate the quantized feature sets $\dot{\mathbf{Z}}$ and $\ddot{\mathbf{Z}}$. We denote the deduplicate set as $\dot{\mathbf{Z}}'=(\dot{\mathbf{z}}'_1,...,\dot{\mathbf{z}}'_t,...,\dot{\mathbf{z}}'_T)^\top$ and $\ddot{\mathbf{Z}}'=(\ddot{\mathbf{z}}'_1,...,\ddot{\mathbf{z}}'_l,...,\ddot{\mathbf{z}}'_L)^\top$.
\paragraph{Code Distribution Re-Targeting.} 
The data distributions of features in the encoder-codebook and decoder-codebook can be very different. 
To illustrate this, we performed kernel density estimations on five randomly selected features trained on the PCN dataset. As shown in Figure~\ref{fig:CD}, a noticeable disparity in data distribution between the two codebooks is evident. 
This poses a challenge when exchanging and merging the quantized information obtained from them.
To address this challenge, a code distribution re-targeting module is devised for merging the two quantized feature sets. 
It consists of two streams: a forward stream from the encoder-codebook to the decoder-codebook and a reverse stream from the decoder-codebook to the encoder-codebook.
\input{ tables/shapenet_part}
In the forward stream, for example, the core of the \textit{re-targeting} module is based on a cascading MLP network. It takes the deduplicated quantized feature set $\dot{\mathbf{Z}}'$ as input and projects them into the decoder-codebook's feature space to align with the target distribution. We denote the re-targeted results as: $\dot{\mathbf{Z}}_\text{r}=(\dot{\mathbf{z}}_r^1,...,\dot{\mathbf{z}}_r^t,...,\dot{\mathbf{z}}_r^T)$. 

\paragraph{Code Merging.} 
To merge the re-targeted code vectors obtained from another codebook, a \textit{code merging} module is designed with three steps. 1) For a re-targeted representation with $\dot{\mathbf{z}}_r^t \in \dot{\mathbf{Z}}_r$, it searches all the codes in the target codebook and identifies the codes with the minimum Euclidean ($\ell_2$) distance , denoted as $\dot{\mathbf{z}}_{s}^i$.
2) the selected codes from decoder-codebook and $\dot{\mathbf{z}}_r^i$ is aggregated in an adaptive manner:
 \begin{equation}
\dot{\mathbf{z}}_\text{res} = \alpha(\dot{\mathbf{z}}_r^{i}, \dot{\mathbf{z}}_s^i)\dot{\mathbf{z}}_r^{i} + (1-\alpha(\dot{\mathbf{z}}_r^{i}, \dot{\mathbf{z}}_s^i))\dot{\mathbf{z}}_s^i,
\end{equation}
where $\alpha(\dot{\mathbf{z}}_r^{i}, \dot{\mathbf{z}}_s^i)$ denotes an adaptive factor, which can be computed as:
\begin{equation}
    \alpha(\dot{\mathbf{z}}_r^{i}, \dot{\mathbf{z}}_s^i) = \frac{\dot{\mathbf{z}}_r^{i} \cdot \dot{\mathbf{z}}_s^i} {|\dot{\mathbf{z}}_r^{i}||\dot{\mathbf{z}}_s^i|}
\end{equation}
This adaptive aggregation controls the degree of the fusion of $\dot{\mathbf{z}}_r^{i}$  and $\dot{\mathbf{z}}_s^i$ in line with their similarity. A higher similarity score suggests the confidence to fuse the two vectors. 
3) 
$\dot{\mathbf{z}}_\text{res}$ is treated as the final quantized vectors for the shape decoder.
\input{./tables/KITTI}
\subsection{Optimization}
\noindent \textbf{Learning on Codebooks. } A contrastive learning based loss is devised to enhance the unique patterns that each code represented for and ensure the codebook coverage for diverse characteristics. 
Specifically, for $\dot{\mathbf{z}}_r^{i} \in \dot{\mathbf{Z}}_\text{r}$, we consider all quantized representations except itself as negative examples. Furthermore, we are dedicated to the reduction of the disparity between the two features, $\dot{\mathbf{z}}_r^i \in \dot{\mathbf{Z}}_r$ and its corresponding $\dot{\mathbf{z}}_s^i$, with utmost precision, thereby ensuring minimal variations among the transformed features.
Based on this, the loss function consists of two parts: an internal loss $\mathcal{L}_\text{internal}$ and an external loss $\mathcal{L}_\text{external}$ as follows:
\begin{equation}
    \mathcal{L}_\text{internal} = \frac{1}{T}(\sum_{i=0}^{T}\sum_{j\neq i}^{T} (1-||\dot{\mathbf{z}}_r^{i}-\dot{\mathbf{z}}_r^{j}||_{2})),
\end{equation}
\begin{equation}
   \mathcal{L}_\text{external} =\frac{1}{T}(\sum_{i=0}^{T}||\dot{\mathbf{z}}_r^{i}-\dot{\mathbf{z}}_s^i||_{2}).
\end{equation}
The internal loss increases the distance between features, promoting a more uniform feature coverage. The external loss aligns the distribution of $\dot{\mathbf{Z}}_\text{r}$ with the decoder-codebook. 
To this end, the loss for learning codebooks can be summarized as follows:
\begin{equation}
    \mathcal{L}_\text{codebook} = \mathcal{L}_\text{internal} + \mathcal{L}_\text{external}.
\end{equation}
\input{ tables/ShapeNet34} 
\noindent \textbf{Learning on Point Cloud Completion.}
We introduce Chamfer Distance (CD) to measure the differences between two point clouds. In detail, for the completed point cloud $\mathcal{P_\text{complete}}$ and its ground truth $\mathcal{P}_\text{GT}$, CD can be defined as follows:
\begin{equation}
\begin{aligned}
     \mathcal{L}_\text{CD}(\mathcal{P}_\text{complete},\mathcal{P}_\text{GT})&=\frac{1}{|\mathcal{P}_\text{complete}|} \sum _{x \in \mathcal{P}_\text{complete}} \min \limits _{y \in \mathcal{P}_\text{GT}} ||x-y||_{2} \\
     &+ \frac{1}{|\mathcal{P}_\text{GT}|} \sum _{y \in \mathcal{P}_\text{GT}} \min \limits _{x \in \mathcal{P}_\text{complete}} ||y-x||_{2}.
\end{aligned}
\end{equation}
Additionally, we adopt a loss function $\mathcal{L}_\text{CD}(\mathcal{P}_\text{coarse},\mathcal{P}_\text{GT}$ to guide the coarse-level completion. Overall the loss function for supervision in this study can be written as: 
\begin{equation}
    \mathcal{L} = \mathcal{L}_\text{CD}(\mathcal{P}_\text{complete},\mathcal{P}_\text{GT}) + \mathcal{L}_\text{CD}(\mathcal{P}_\text{coarse},\mathcal{P}_\text{GT}) + \mathcal{L}_\text{codebook}.
\end{equation}

\section{Experiments \& Discussions}
\subsection{Experimental Settings}
\paragraph{Datasets.} To demonstrate the effectiveness of the proposed DC-PCN, we conducted the evaluation on three widely used datasets. PCN,  ShapeNet\_Part, and ShapeNet34. 
\begin{itemize}
\item \textbf{PCN} \cite{yuan2018pcn} consists of 28,974 shapes for training and 1,200 shapes for testing from 8 categories. The ground-truth point cloud was created by random sampling 16,384 points from each shape. Similarly, partial point clouds are produced by sampling incomplete point clouds from the corresponding shapes. 
\item \textbf{ShapeNet\_Part} contains 14,473 shapes and was split into 11,705 shapes for training and 2,768 for testing. These shapes belong to 13 categories. All shapes were normalized to the range of [-1,1] and centered at the origin for consistency. 
\item \textbf{ShapeNet34.} To evaluate the method's generalizability, following the experimental setup in \cite{yu2023adapointr}, we partitioned the ShapeNet dataset into two distinct subsets: 34 visible categories and 21 unseen categories. Within the visible categories, a subset of 100 objects per category was randomly selected to comprise the visible test set, ensuring the remaining objects were utilized for training purposes.For the unseen categories, a comprehensive collection of 2,305 objects was designated as the test set. 
\end{itemize}

\paragraph{Evaluation Metrics.} Following the existing practices, the evaluation metrics adopted in this study includes Chamfer Distance (CD) and F-score@\%1~\cite{tatarchenko2019single}  between the predicted complete shape and the ground truth. 
\paragraph{Implementation Details.}
All experiments were conducted utilizing two RTX 2080Ti GPUs

\subsection{Performance on PCN Dataset}
For the evaluation on PCN, we follow the standard protocol and evaluation metric ($\ell_1$ chamfer distance) and F-Score as in existing studies \cite{yang2018foldingnet,groueix2018papier,yuan2018pcn,tchapmi2019topnet,mitra2013symmetry,xie2020grnet,wang2020cascaded,wen2021pmp,zhang2020detail,xiang2021snowflakenet,tang2022lake,zhou2022seedformer,yu2021pointr,yu2023adapointr,zhu2023svdformer,zhu2024advancements,wu2024fsc}.The results are listed in Table \ref{tab:PCN}. 
It is evident that our DC-PCN achieves the best CD-$\ell_1$ and F-Score 0.850 in a majority of categories, indicating the superiority of DC-PCN compared with other methods. 
Notably, our DC-PCN improves the average CD-$\ell_1$ by 0.07, compared with the second-best method - AdaPoinTR. 

\begin{figure*}
    \centering
    \includegraphics[width=\linewidth]{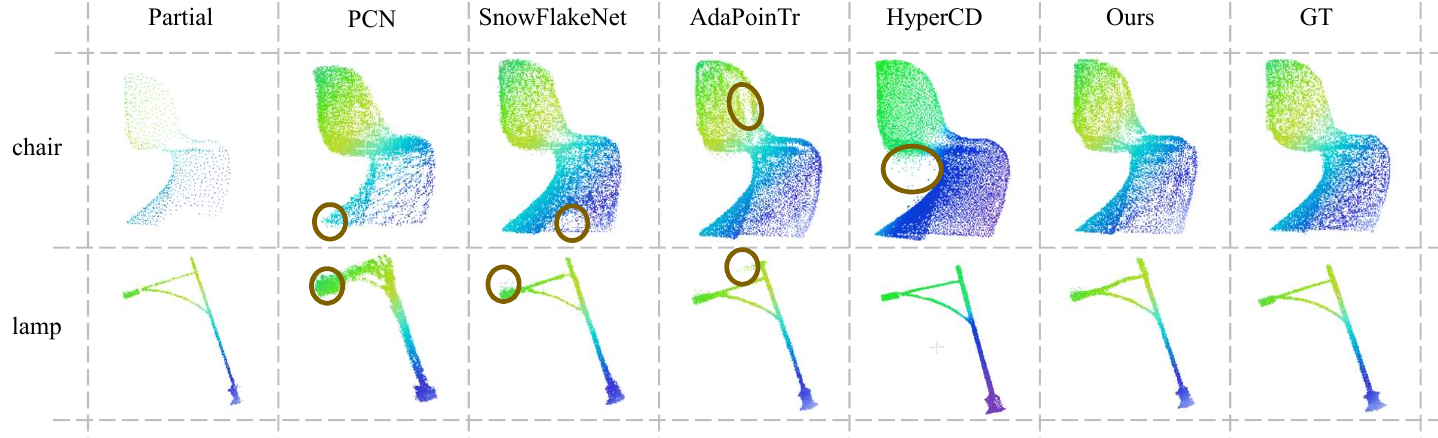}
    \caption{Visualization of the results on  PCN. The results of our method show higher-level noise suppression and have refined geometric structure compared to those obtained from existing methods.}
    \label{fig:compareInPCN}
\end{figure*}

Qualitative examples are visualized in Figure \ref{fig:compareInPCN}, where we contrast our outcomes with those of PCN \cite{yuan2018pcn}, SnowFlakeNet \cite{xiang2021snowflakenet}, AdaPoinTR \cite{yu2023adapointr} and HyperCD \cite{zhu2024advancements} across two distinct categories: chair and lamp. 
These examples demonstrate our DC-PCN's completion results exhibit exceptional noise suppression and high-quality geometric details. 
\subsection{Performance on ShapeNet\_Part Dataset}
ShapeNet\_Part assesses the performance in a multi-category and sparse context. 
Table \ref{tab:shapeNet_part} lists the quantitative results for all methods, measured by the CD-$\ell_2$ metric. As HyperCD \cite{zhu2024advancements}, SVDFormer \cite{zhu2023svdformer}, FSC \cite{wu2024fsc} have not evaluated their method in this dataset, we compare our method with the rest point cloud completion methods in ShapeNet\_Part. It can be observed that our method achieves the best CD-$\ell_2$ in the majority of categories, as well as a best average CD-$\ell_2$ 5.59. 
In contrast, AdaPoinTR \cite{yu2023adapointr} and PoinTR \cite{yu2021pointr} achieved 6.10 and 6.26, respectively, demonstrating the significant improvement offered by our method.
\subsection{Performance on ShapeNet34 Dataset}
Table \ref{tab:shapeNet34} outlines the CD-$\ell_2$ metric achieved by our method on both the seen categories and the unseen classes for ShapeNet34. 
It is anticipated that the average performance on the unseen categories would be inferior to that of the seen categories due to the inherent new knowledge of these objects. 
Despite the substantial variations in shape geometry between the training and testing data, our DC-PCN demonstrates remarkable resilience, achieving the best result on the case of simple, medium and hard, moreover achieving the best CD-$\ell_2$ among all the methods. Moreover, regarding the F-Score, our method achieves comparable performance with the state-of-the-art performance. This demonstrates the generalizability of our DC-PCN.
\subsection{Performance on KITTI Dataset}
To further demonstrate the generalizability of DC-PCN, we conducted evaluation on the KITTI dataset as shown in Table \ref{tab:KITTI}, which contains real-world data with diverse object appearances. with a pre-trained model on PCN cars. Our model achieves an MMD metric of 0.373, represent
ing an improvement of 0.019 compared to the current best model- AdaPointr with an MMD of 0.392. Our DC-PCN shows the capabilities for addressing real-world point clouds.
\subsection{Ablation Study}
\begin{figure*}[h]
    \centering
    \includegraphics[width=\linewidth]{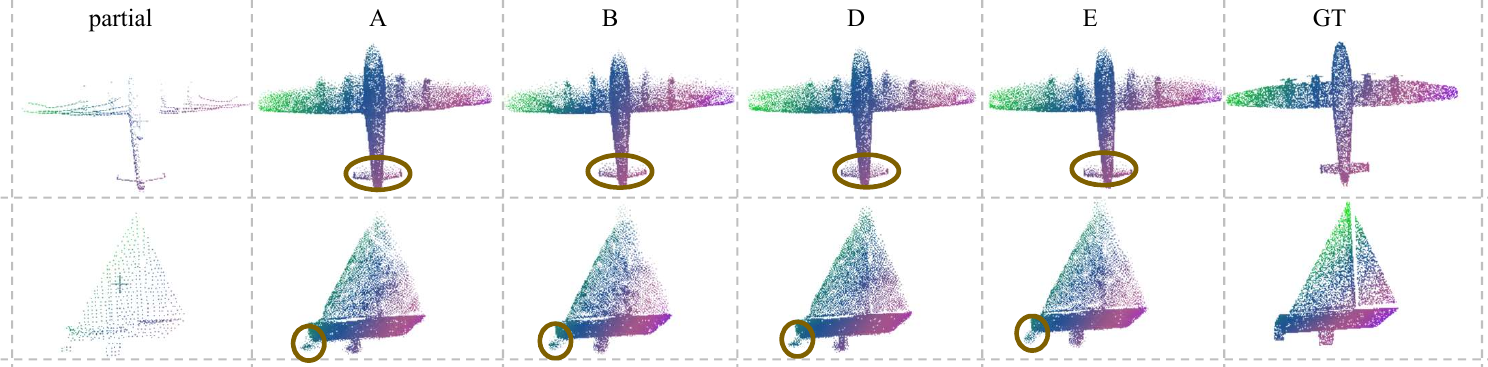}
    \caption{Visualization of the results on ablation settings A, B, D and E. Our method exhibits  superiority when dealing with fine-grained completion structures.}
    \label{fig:abv}
\end{figure*}

To evaluate the effectiveness of each proposed mechanism in our DC-PCN, we conducted ablation studies with the PCN benchmark. The results are outlined in Table \ref{tab:ablation}. Further more, we visualize some of them in Figure \ref{fig:abv}
\paragraph{Dual-Codebook.} We evaluated the effectiveness of the dual-codebook design, which represents the discrete feature distribution at shallow-level and deep-level. 
In detail, we compare the performance of Method A without the encoder-codebook and Method B with the encoder-codebook. 
It can be observed that the encoder-codebook (EC) improves the CD-$\ell_1$ and F-score@1 by 0.05 and 0.005. 
In parallel, comparing Method C with the decoder-codebook (DC)  and Method A without the decoder-codebook, the metrics indicate an  improvement 0.01 in terms of CD-$\ell_1$. 
In parallel, comparing Method D with the dual-codebook deign and Method A without it, the metric indicates an improvement 0.06 regarding CD-$\ell_1$.
To this end, the experimental results demonstrate the necessity of the quantization in the feature space, highlighting the important roles of our encoder-codebook and decoder-codebook. 
\input{tables/ablation}
\paragraph{Quantized Information Exchanging.} 
Further more, in order to verify the \textit{quantized information exchanging} mechanism (QIE), we consider the ablation setting as Method D, which lacks this mechanism, and Model E, which is with it. The results show an improvement of 0.01 in CD-$\ell_1$ and 0.002 in F-Score@1, respectively. It forcefully demonstrates the effectiveness of the fusion mechanism. 
Further more, in order to verify the \textit{quantized information exchanging} mechanism, we consider the ablation setting as Method D, which lacks this mechanism, and Model E, which is with it. The results show an improvement of 0.01 in CD-$\ell_1$ and 0.002 in F-Score@1, respectively. It forcefully demonstrates the effectiveness of the fusion mechanism. 
\paragraph{A Shared Codebook}
In order to further verify the effectiveness of the decoder codebook and the fusion mechanism, we set up Group F experiments to make a codebook simultaneously learn the discrete distributions of shallow features and deep features. Judging from the experimental results, the completion effect was significantly affected. Compared with the situation without the codebook, the CD-$\ell_1$ loss in the experimental results increased by 0.15.
\section{Limitations}
The major limitation of DC-PCN is its reliance on the codebook design, which involves a series of hyperparameters. For example, the optimal size of the codebooks varies across different datasets and has the impacts on performance.
The sizes of codebooks in the experiments are all the optimal values obtained by changing different parameters. Future work should focus on exploring an adaptive codebook production mechanism in a data-driven manner.

\section{Conclusion}
We presented a point cloud completion network - DC-PCN for creating a consistent representation for sampled point clouds originating from the same 3D surface through a vector discretization strategy. 
DC-PCN introduces a dual-codebook design to quantize point cloud representations from a multi-level perspective with an encoder-codebook and a decoder-codebook. An information exchange mechanism further enhances the information flow between the two codebooks. 
Comprehensive experiments demonstrate the state-of-the-art performance of our method.

\bibliography{aaai25}
\end{document}

%% file: tables/PCN.tex
\begin{table*}[htbp]
   
    \centering
     \setlength{\tabcolsep}{1mm}{
    \begin{tabular}{l|cc|cccccccc}
    \toprule
        \parbox{4cm}{Methods} & {Average} &{F-Score@1}  & {Plane } & {Cabinet} & {Car} & {Chair} & {Lamp} & {Sofa} &{Table} & {Watercraft}  \\ \midrule
        {FoldingNet \cite{yang2018foldingnet}} &14.31 & - &9.49&15.80&12.61&15.55&16.41&15.97& 13.65&14.99\\ 
        {AtlasNet \cite{groueix2018papier}} &10.85 & - &6.37&11.94&10.10&12.06&12.37&12.99&10.33&10.61\\
        {PCN \cite{yuan2018pcn}} &9.64 &0.695&5.50&22.70&10.63&8.70&11.00&11.34&11.68&8.59\\ 
        {TopNet \cite{tchapmi2019topnet}} &12.15&-&7.61&13.31&10.90&13.82&14.44&14.78&11.22&11.12\\ 
        {MSN \cite{liu2020morphing}} &10.00&-&5.60&11.90&10.30&10.20&10.70&11.60&9.60&9.90\\
        {GRNet \cite{xie2020grnet}} &8.83&0.708&6.45&10.37&9.45&9.41&7.96&10.51&8.44&8.04\\ 
        {CRN \cite{wang2020cascaded}} &8.51&0.652&4.79&9.97&8.31&9.49&8.94&10.69&7.81&8.05\\
        {NSFA \cite{zhang2020detail}} &8.06&0.734&4.76&10.18&8.63&8.53&7.03&10.53&7.35&7.48\\ 
        {SnowFlake \cite{xiang2021snowflakenet}} &7.21&0.801&4.29&9.16&8.08&7.89&6.07&9.23&6.55&6.40\\ 
         {Pointr \cite{yu2021pointr}} &8.38&0.745&4.75&10.47&8.68&9.39&7.75&10.93&7.78&7.29\\ 
       {PMP-Net \cite{wen2021pmp}} &8.73&0.781&5.65&11.24&9.64&9.51&6.95&10.83&8.72&7.25\\ 
        {LAKeNet \cite{tang2022lake}} &7.23&-&4.17&9.78&8.56&7.45&5.88&9.39&6.43&5.98\\ 
        {SeedFormer \cite{zhou2022seedformer}} &6.74&0.818&3.85&9.05&8.06&7.06&\underline{5.21}&8.85&6.05&5.85\\ 
        {AdaPointr \cite{yu2023adapointr}} &\underline{6.53}&\underline{0.845}&3.68&\underline{8.82}&\underline{7.47}&6.85&5.47&\underline{8.35}&\underline{5.80}&5.76\\ 
        {SVDFormer \cite{zhu2023svdformer}} &6.54&0.841&\textbf{3.62}&8.79&\textbf{7.46}&6.91&5.33&8.49&5.90&5.83\\ 
         {HyperCD \cite{zhu2024advancements}} &6.54&-&3.72&\textbf{8.71}&7.79&\underline{6.83}&\textbf{5.11}&8.61&5.82&5.76\\ 
        {FSC \cite{wu2024fsc}} &7.02 &-&4.07&9.12&8.1&7.21&5.88&9.30&6.26&\textbf{5.25}\\ 
        
        \midrule
        {Ours} &\textbf{6.46}&\textbf{0.850} &\underline{3.65}&\underline{8.75} &7.48&\textbf{6.71}&5.35&\textbf{8.28}&\textbf{5.76}&\underline{5.71}\\ 
        \bottomrule
    \end{tabular}}
    
     \caption{Quantitative comparisons on the PCN dataset using CD-$\ell_1$ $\downarrow$ (scaled by $10^3$) and F-Score.}
    \label{tab:PCN}
\end{table*}

%% file: tables/shapenet_part.tex
\begin{table*}[htbp]
  
	\centering
         \setlength{\tabcolsep}{1mm}{
	\begin{tabular}{l|ccccccccccccc|c}
		\toprule
		Methods & {Plane} & {Bag} & {Cap} & {Car} & {Chair} &{ GTar} & {Lamp} & {PC} & {Moto} & {Mug} & {Pistol} & {Skate} & {Table} & {Avg} \\ \midrule
		{FoldingNet \cite{yang2018foldingnet}}&11.9&48.3&52.0&30.2&21.5&6.0&46.5&14.4&25.6&41.9&18.0&17.4&26.3&27.7\\
		{PCN \cite{yuan2018pcn}}&11.4&44.9&51.4&30.2&20.6&5.8&42.1&13.6&24.5&42.6&17.3&14.2&24.6&26.4\\
		{TopNet \cite{tchapmi2019topnet}}&12.7&47.5&52.9&32.8&22.2&6.0&43.1&16.0&28.0&42.9&17.9&15.6&24.0&27.8\\
		{PF-Net \cite{huang2020pf}}&13.1&24.7&48.3&14.0&12.7&12.8&61.1&9.0&16.6&22.7&23.4&19.7&18.4&22.8\\
		{GRNet \cite{xie2020grnet}}&11.6&25.5&24.8&25.5&17.9&6.7&22.4&16.5&20.7&35.4&13.1&12.8&18.1&19.3\\
		{MSN \cite{liu2020morphing}}&8.2&24.3&25.0&18.8&12.8&6.4&30.4&8.9&20.2&18.3&13.6&9.1&14.2&16.2\\
		{SeedFormer \cite{zhou2022seedformer}}&4.5&10.6&9.5&11.3&7.0&2.3&10.2&7.0&\underline{9.1}&16.2&5.8&5.4&7.8&8.2\\
		{PoinTr \cite{yu2021pointr}}      &3.8&\underline{8.3}&\textbf{5.5}&7.2&\underline{4.5}&2.6&12.7&\textbf{3.4}&10.3&\textbf{6.7}&5.5&\underline{3.9}&\underline{5.0}&\underline{6.1}\\
		{AdaPoinTr \cite{yu2023adapointr}}&\underline{3.6}&8.8&\underline{6.0}&\underline{8.5}&5.0&\textbf{2.1}&\textbf{8.1}&4.7&9.1&10.4&\underline{5.3}&3.9&5.8&6.3 \\
  \midrule
		Ours&\textbf{3.4}&\textbf{7.4}&6.3&\textbf{6.8}&\textbf{4.3}&\underline{2.1}&\underline{9.5}&\underline{4.1}&\textbf{6.9}&\underline{8.5}&\textbf{4.9}&\textbf{3.8}&\textbf{4.8}&\textbf{5.6}                         
              \\ \bottomrule
	\end{tabular}}
	\caption{Quantitative comparisions on the ShapeNet\_Part dataset using CD-$\ell_2$ $\downarrow$ (scaled by $10^4$).}
	\label{tab:shapeNet_part}
\end{table*}

%% file: tables/kitti.tex
\begin{table*}[htbp]
    \centering
    \setlength{\tabcolsep}{1mm}{
    \begin{tabular}{l|cccccccccc|c}
     \toprule {CD-$\ell_2$\(\times{10^3} \)}& SeedFormer \cite{zhou2022seedformer} & PoinTr \cite{yu2021pointr} & AdaPoinTr \cite{yu2023adapointr} & Our \\

     \midrule {MMD}& 0.516 & 0.526 & 0.392 & 0.373 \\
     \bottomrule
    \end{tabular}}
    \caption{Quantitative comparisons on the KITTI dataset using MMD $\downarrow$}
    \label{tab:KITTI}
\end{table*}

%% file: tables/shapenet34.tex
\begin{table*}
	\centering
     \setlength{\tabcolsep}{1mm}{
	\begin{tabular}{l|ccccc|ccccc}
		\toprule
		\multirow{2}{*}{Methods} & \multicolumn{5}{|c|}{34 seen categories}& \multicolumn{5}{c}{21 unseen categories}\\ \cmidrule(lr){2-6} \cmidrule(lr){7-11}
		    & CD-S& CD-M& CD-H& CD-$\ell_2$& F-Score@1& CD-S& CD-M& CD-H& CD-$\ell_2$&F-Score@1 \\ \midrule
		{FoldingNet \cite{yang2018foldingnet}}& 1.86& 1.81& 3.38& 2.35& 0.139& 2.76& 2.74& 5.36& 3.62& 0.095      \\
		\parbox{4cm}{PCN \cite{yuan2018pcn}}& 1.87& 1.81& 2.97& 2.22& 0.154& 3.17& 3.08& 5.29& 3.85& 0.101      \\
		{TopNet \cite{tchapmi2019topnet}}& 1.77& 1.61& 3.54& 2.31& 0.171& 2.62& 2.43& 5.44& 3.50& 0.121      \\
		{PFNet \cite{huang2020pf}}& 3.16& 3.19& 7.71& 4.68& 0.347& 5.29& 5.87& 13.33& 8.16& 0.322      \\
		{GRNet \cite{xie2020grnet}}& 1.26& 1.39& 2.57& 1.74& 0.251& 1.85& 2.25& 4.87& 2.99& 0.216      \\
		{SnowflakeNet \cite{xiang2021snowflakenet}}& 0.60& 0.86& 1.50& 0.99& 0.42& 0.88& 1.46& 2.92& 1.75& 0.388      \\
		{SeedFormer \cite{zhou2022seedformer}}& \underline{0.48}& 0.70& 1.30& 0.83& 0.452& \underline{0.61}& 1.07& 2.35& 1.34& 0.402      \\
		{PoinTr  \cite{yu2021pointr}}& 0.76& 1.05& 1.88& 1.23& 0.421& 1.04& 1.67& 3.44& 2.05& 0.384      \\
		{AdaPoinTr \cite{yu2023adapointr}}& \underline{0.48}& \underline{0.63}& \underline{1.07}& \underline{0.73}& \underline{0.47}& \underline{0.61}& \underline{0.96}& \underline{2.11}& \underline{1.23}& 0.416    \\
        {SVDFormer \cite{zhu2023svdformer} }& \textbf{0.46}& 0.65& 1.13& 0.75& 0.457& \underline{0.61}& 1.05& 2.19& 1.28& \underline{0.427}\\
        {HyperCD \cite{zhu2024advancements}}& \textbf{0.46}& 0.67& 1.24& 0.79& 0.459& \textbf{0.58}& 1.03&2.24& 1.31&\textbf{ 0.428}     
  \\ \midrule
		{Ours}&0.49&\textbf{0.56}&\textbf{0.99}&\textbf{0.68}&\textbf{0.48} &\textbf{0.58}&\textbf{0.87}&\textbf{1.94}&\textbf{1.13}&0.422\\ \bottomrule
	\end{tabular}}
    
     \caption{
    Quantitative comparisions on the ShapeNet-34 dataset. 
    The evaluation were conducted with three  difficulty levels: CD-S, CD-M, and CD-H. Evaluation metrics are CD-$\ell_2$ (scaled by $10^3$) $\downarrow$ and F-Score@1\% $\uparrow$.}
	\label{tab:shapeNet34}
\end{table*}

%% file: tables/ablation.tex
\begin{table}
    \centering
   \setlength{\tabcolsep}{1mm}{
    \begin{tabular}{c|ccc|ccc}
    \toprule
        Mtd & EC & DC  & QIE & CD-$\ell_1$ & CD-$\ell_2$ & F-score@1 \\ 
        \midrule
        A & ~ & ~ & ~ &6.53&0.194&0.845\\ 
        B & \checkmark & ~  & ~ &6.48&0.193&0.850\\ 
        C & ~ & \checkmark & ~ &6.52&0.194&0.843\\ 
        D & \checkmark & \checkmark & ~ &6.47&0.192&0.848 \\ 
        E & \checkmark & \checkmark  & \checkmark &6.46&0.192&0.850\\ 
        F & \multicolumn{2}{c}{shared codebook} & ~ &6.685&0.200&0.837\\  
    \bottomrule
    \end{tabular}}
    
        \caption{
Ablation studies conducted on PCN with the evaluation metrics CD-$\ell_1$ $\downarrow$ (scaled by $10^3$), CD-$\ell_2$ $\downarrow$ (scaled by $10^3$), and F-Score@1 $\uparrow$ metrics.
}
	\label{tab:ablation}
\end{table}

%% file: aaai25.bbl
\begin{thebibliography}{36}
\providecommand{\natexlab}[1]{#1}

\bibitem[{Choy et~al.(2016)Choy, Xu, Gwak, Chen, and Savarese}]{choy20163d}
Choy, C.~B.; Xu, D.; Gwak, J.; Chen, K.; and Savarese, S. 2016.
\newblock 3{D}-{R}2{N}2: A Unified Approach for Single and Multi-View 3D Object Reconstruction.
\newblock In \emph{European Conference on Computer Vision}, 628--644. Springer.

\bibitem[{Cui et~al.(2023)Cui, Qiu, Anwar, Liu, Xing, Zhang, and Barnes}]{cui2023p2c}
Cui, R.; Qiu, S.; Anwar, S.; Liu, J.; Xing, C.; Zhang, J.; and Barnes, N. 2023.
\newblock P2c: Self-Supervised Point Cloud Completion from Single Partial Clouds.
\newblock In \emph{IEEE/CVF International Conference on Computer Vision}, 14351--14360.

\bibitem[{Dai, Ruizhongtai~Qi, and Nie{\ss}ner(2017)}]{dai2017shape}
Dai, A.; Ruizhongtai~Qi, C.; and Nie{\ss}ner, M. 2017.
\newblock Shape Completion Using 3D-Encoder-Predictor Cnns and Shape Synthesis.
\newblock In \emph{IEEE Conference on Computer Vision and Pattern Recognition}, 5868--5877.

\bibitem[{Fei et~al.(2022{\natexlab{a}})Fei, Yang, Chen, Li, Li, Ma, Hu, and Ma}]{fei2022comprehensive}
Fei, B.; Yang, W.; Chen, W.-M.; Li, Z.; Li, Y.; Ma, T.; Hu, X.; and Ma, L. 2022{\natexlab{a}}.
\newblock Comprehensive Review of Deep Learning-Based 3D Point Cloud Completion Processing and Analysis.
\newblock \emph{IEEE Transactions on Intelligent Transportation Systems}.

\bibitem[{Fei et~al.(2022{\natexlab{b}})Fei, Yang, Chen, and Ma}]{fei2022vq}
Fei, B.; Yang, W.; Chen, W.-M.; and Ma, L. 2022{\natexlab{b}}.
\newblock VQ-DcTr: Vector-Quantized Autoencoder with Dual-Channel Transformer Points Splitting for 3D Point Cloud Completion.
\newblock In \emph{ACM International Conference on Multimedia}, 4769--4778.

\bibitem[{Girdhar et~al.(2016)Girdhar, Fouhey, Rodriguez, and Gupta}]{girdhar2016learning}
Girdhar, R.; Fouhey, D.~F.; Rodriguez, M.; and Gupta, A. 2016.
\newblock Learning a Predictable and Generative Vector Pepresentation for Objects.
\newblock In \emph{European Conference on Computer Vision}, 484--499. Springer.

\bibitem[{Groueix et~al.(2018)Groueix, Fisher, Kim, Russell, and Aubry}]{groueix2018papier}
Groueix, T.; Fisher, M.; Kim, V.~G.; Russell, B.~C.; and Aubry, M. 2018.
\newblock A Papier-M{\^a}ch{\'e} Approach to Learning 3D Surface Generation.
\newblock In \emph{IEEE Conference on Computer Vision and Pattern Recognition}, 216--224.

\bibitem[{Guo et~al.(2021)Guo, Cai, Liu, Mu, Martin, and Hu}]{guo2021pct}
Guo, M.-H.; Cai, J.-X.; Liu, Z.-N.; Mu, T.-J.; Martin, R.~R.; and Hu, S.-M. 2021.
\newblock PCT: Point Cloud Transformer.
\newblock \emph{Computational Visual Media}, 7: 187--199.

\bibitem[{Han et~al.(2017)Han, Li, Huang, Kalogerakis, and Yu}]{han2017high}
Han, X.; Li, Z.; Huang, H.; Kalogerakis, E.; and Yu, Y. 2017.
\newblock High-Resolution Shape Completion Using Deep Neural Networks for Global Structure and Local Geometry Inference.
\newblock In \emph{IEEE International Conference on Computer Vision}, 85--93.

\bibitem[{Huang et~al.(2020)Huang, Yu, Xu, Ni, and Le}]{huang2020pf}
Huang, Z.; Yu, Y.; Xu, J.; Ni, F.; and Le, X. 2020.
\newblock Pf-Net: Point Fractal Network for 3D Point Cloud Completion.
\newblock In \emph{IEEE/CVF Conference on Computer Vision and Pattern Recognition}, 7662--7670.

\bibitem[{Liu et~al.(2020)Liu, Sheng, Yang, Shao, and Hu}]{liu2020morphing}
Liu, M.; Sheng, L.; Yang, S.; Shao, J.; and Hu, S.-M. 2020.
\newblock Morphing and Sampling Network for Dense Point Cloud Completion.
\newblock In \emph{AAAI Conference on Artificial Intelligence}, volume~34, 11596--11603.

\bibitem[{Mitra et~al.(2013)Mitra, Pauly, Wand, and Ceylan}]{mitra2013symmetry}
Mitra, N.~J.; Pauly, M.; Wand, M.; and Ceylan, D. 2013.
\newblock Symmetry in 3D Geometry: Extraction and Applications.
\newblock In \emph{Computer Graphics Forum}, volume~32, 1--23. Wiley Online Library.

\bibitem[{Mittal et~al.(2022)Mittal, Cheng, Singh, and Tulsiani}]{mittal2022autosdf}
Mittal, P.; Cheng, Y.-C.; Singh, M.; and Tulsiani, S. 2022.
\newblock Autosdf: Shape Priors for 3D Completion, Reconstruction and Generation.
\newblock In \emph{IEEE/CVF Conference on Computer Vision and Pattern Recognition}, 306--315.

\bibitem[{Mo et~al.(2025)Mo, Hu, Long, Yuan, and Wang}]{mo2025motion}
Mo, C.; Hu, K.; Long, C.; Yuan, D.; and Wang, Z. 2025.
\newblock Motion Keyframe Interpolation for Any Human Skeleton via Temporally Consistent Point Cloud Sampling and Reconstruction.
\newblock In \emph{European Conference on Computer Vision}, 159--175. Springer.

\bibitem[{Pan et~al.(2021)Pan, Xia, Song, Li, and Huang}]{pan20213d}
Pan, X.; Xia, Z.; Song, S.; Li, L.~E.; and Huang, G. 2021.
\newblock 3D Object Detection with Pointformer.
\newblock In \emph{IEEE/CVF Conference on Computer Vision and Pattern Recognition}, 7463--7472.

\bibitem[{Qi et~al.(2017)Qi, Su, Mo, and Guibas}]{qi2017pointnet}
Qi, C.~R.; Su, H.; Mo, K.; and Guibas, L.~J. 2017.
\newblock Pointnet: Deep Learning on Point Sets for 3D Classification and Segmentation.
\newblock In \emph{IEEE Conference on Computer Vision and Pattern Recognition}, 652--660.

\bibitem[{Tang et~al.(2022)Tang, Gong, Yi, Xie, and Ma}]{tang2022lake}
Tang, J.; Gong, Z.; Yi, R.; Xie, Y.; and Ma, L. 2022.
\newblock Lake-Net: Topology-Aware Point Cloud Completion by Localizing Aligned Keypoints.
\newblock In \emph{IEEE/CVF Conference on Computer Vision and Pattern Recognition}, 1726--1735.

\bibitem[{Tatarchenko et~al.(2019)Tatarchenko, Richter, Ranftl, Li, Koltun, and Brox}]{tatarchenko2019single}
Tatarchenko, M.; Richter, S.~R.; Ranftl, R.; Li, Z.; Koltun, V.; and Brox, T. 2019.
\newblock What Do Single-View 3D Reconstruction Networks Learn?
\newblock In \emph{IEEE/CVF Conference on Computer Vision and Pattern Recognition}, 3405--3414.

\bibitem[{Tchapmi et~al.(2019)Tchapmi, Kosaraju, Rezatofighi, Reid, and Savarese}]{tchapmi2019topnet}
Tchapmi, L.~P.; Kosaraju, V.; Rezatofighi, H.; Reid, I.; and Savarese, S. 2019.
\newblock Topnet: Structural Point Cloud Decoder.
\newblock In \emph{IEEE/CVF Conference on Computer Vision and Pattern Recognition}, 383--392.

\bibitem[{Van Den~Oord, Vinyals et~al.(2017)}]{van2017neural}
Van Den~Oord, A.; Vinyals, O.; et~al. 2017.
\newblock Neural Discrete Representation Learning.
\newblock \emph{Advances in Neural Information Processing Systems}, 30.

\bibitem[{Wang, Ang~Jr, and Lee(2020)}]{wang2020cascaded}
Wang, X.; Ang~Jr, M.~H.; and Lee, G.~H. 2020.
\newblock Cascaded Refinement Network For Point Cloud Completion.
\newblock In \emph{IEEE/CVF Conference on Computer Vision and Pattern Recognition}, 790--799.

\bibitem[{Wen et~al.(2021)Wen, Xiang, Han, Cao, Wan, Zheng, and Liu}]{wen2021pmp}
Wen, X.; Xiang, P.; Han, Z.; Cao, Y.-P.; Wan, P.; Zheng, W.; and Liu, Y.-S. 2021.
\newblock Pmp-Net: Point Cloud Completion by Learning Multi-Step Point Moving Paths.
\newblock In \emph{IEEE/CVF Conference on Computer Vision and Pattern Recognition}, 7443--7452.

\bibitem[{Wu et~al.(2024)Wu, Wu, Luan, Bai, Lai, and Yuan}]{wu2024fsc}
Wu, X.; Wu, X.; Luan, T.; Bai, Y.; Lai, Z.; and Yuan, J. 2024.
\newblock FSC: Few-Point Shape Completion.
\newblock In \emph{IEEE/CVF Conference on Computer Vision and Pattern Recognition}, 26077--26087.

\bibitem[{Wu et~al.(2015)Wu, Song, Khosla, Yu, Zhang, Tang, and Xiao}]{wu20153d}
Wu, Z.; Song, S.; Khosla, A.; Yu, F.; Zhang, L.; Tang, X.; and Xiao, J. 2015.
\newblock 3D Shapenets: A Deep Representation for Volumetric Shapes.
\newblock In \emph{IEEE Conference on Computer Vision and Pattern Recognition}, 1912--1920.

\bibitem[{Xiang et~al.(2021)Xiang, Wen, Liu, Cao, Wan, Zheng, and Han}]{xiang2021snowflakenet}
Xiang, P.; Wen, X.; Liu, Y.-S.; Cao, Y.-P.; Wan, P.; Zheng, W.; and Han, Z. 2021.
\newblock Snowflakenet: Point Cloud Completion by Snowflake Point Deconvolution with Skip-Transformer.
\newblock In \emph{IEEE/CVF International Conference on Computer Vision}, 5499--5509.

\bibitem[{Xie et~al.(2020)Xie, Yao, Zhou, Mao, Zhang, and Sun}]{xie2020grnet}
Xie, H.; Yao, H.; Zhou, S.; Mao, J.; Zhang, S.; and Sun, W. 2020.
\newblock Grnet: Gridding Residual Network for Dense Point Cloud Completion.
\newblock In \emph{European Conference on Computer Vision}, 365--381. Springer.

\bibitem[{Yan et~al.(2022)Yan, Lin, Mitra, Lischinski, Cohen-Or, and Huang}]{yan2022shapeformer}
Yan, X.; Lin, L.; Mitra, N.~J.; Lischinski, D.; Cohen-Or, D.; and Huang, H. 2022.
\newblock Shapeformer: Transformer-Based Shape Completion Via Sparse Representation.
\newblock In \emph{IEEE/CVF Conference on Computer Vision and Pattern Recognition}, 6239--6249.

\bibitem[{Yang et~al.(2018)Yang, Feng, Shen, and Tian}]{yang2018foldingnet}
Yang, Y.; Feng, C.; Shen, Y.; and Tian, D. 2018.
\newblock Foldingnet: Point Cloud Auto-Encoder Via Deep Grid Deformation.
\newblock In \emph{IEEE Conference on Computer Vision and Pattern Recognition}, 206--215.

\bibitem[{Yu et~al.(2021)Yu, Rao, Wang, Liu, Lu, and Zhou}]{yu2021pointr}
Yu, X.; Rao, Y.; Wang, Z.; Liu, Z.; Lu, J.; and Zhou, J. 2021.
\newblock Pointr: Diverse Point Cloud Completion with Geometry-Aware Transformers.
\newblock In \emph{IEEE/CVF International Conference on Computer Vision}, 12498--12507.

\bibitem[{Yu et~al.(2023)Yu, Rao, Wang, Lu, and Zhou}]{yu2023adapointr}
Yu, X.; Rao, Y.; Wang, Z.; Lu, J.; and Zhou, J. 2023.
\newblock AdaPoinTr: Diverse Point Cloud Completion With Adaptive Geometry-Aware Transformers.
\newblock \emph{ArXiv Preprint ArXiv:2301.04545}.

\bibitem[{Yuan et~al.(2018)Yuan, Khot, Held, Mertz, and Hebert}]{yuan2018pcn}
Yuan, W.; Khot, T.; Held, D.; Mertz, C.; and Hebert, M. 2018.
\newblock PCN: Point Completion Network.
\newblock In \emph{International Conference on 3D Vision}, 728--737. IEEE.

\bibitem[{Zhang, Yan, and Xiao(2020)}]{zhang2020detail}
Zhang, W.; Yan, Q.; and Xiao, C. 2020.
\newblock Detail Preserved Point Cloud Completion Via Separated Feature Aggregation.
\newblock In \emph{European Conference on Computer Vision}, 512--528. Springer.

\bibitem[{Zhao et~al.(2021)Zhao, Jiang, Jia, Torr, and Koltun}]{zhao2021point}
Zhao, H.; Jiang, L.; Jia, J.; Torr, P.~H.; and Koltun, V. 2021.
\newblock Point Transformer.
\newblock In \emph{IEEE/CVF International Conference on Computer Vision}, 16259--16268.

\bibitem[{Zhou et~al.(2022)Zhou, Cao, Chu, Zhu, Lu, Tai, and Wang}]{zhou2022seedformer}
Zhou, H.; Cao, Y.; Chu, W.; Zhu, J.; Lu, T.; Tai, Y.; and Wang, C. 2022.
\newblock Seedformer: Patch Seeds Based Point Cloud Completion with Upsample Transformer.
\newblock In \emph{European Conference on Computer Vision}, 416--432. Springer.

\bibitem[{Zhu, Fan, and Weng(2024)}]{zhu2024advancements}
Zhu, Q.; Fan, L.; and Weng, N. 2024.
\newblock Advancements in Point Cloud Data Augmentation for Deep Learning: A Survey.
\newblock \emph{Pattern Recognition}, 110532.

\bibitem[{Zhu et~al.(2023)Zhu, Chen, He, Wang, Qin, and Wei}]{zhu2023svdformer}
Zhu, Z.; Chen, H.; He, X.; Wang, W.; Qin, J.; and Wei, M. 2023.
\newblock Svdformer: Complementing Point Cloud Via Self-View Augmentation and Self-Structure Dual-Generator.
\newblock In \emph{IEEE/CVF International Conference on Computer Vision}, 14508--14518.

\end{thebibliography}
